%% file: arxiv.tex
\documentclass[sigconf]{aamas} 

\usepackage{balance} 

\usepackage{amsthm,amsmath,amssymb}

\usepackage{bm}
\usepackage{mathrsfs}
\usepackage{hyperref}
\usepackage{url}
\usepackage{graphicx}
\usepackage{pifont}
\usepackage{multirow}
\usepackage{hyperref}
\hypersetup{
	colorlinks=true,
	linkcolor=cyan,
	filecolor=blue,      
	urlcolor=red,
	citecolor=green,
}

\setcopyright{ifaamas}
\acmConference[AAMAS '23]{Proc.\@ of the 22nd International Conference
on Autonomous Agents and Multiagent Systems (AAMAS 2023)}{May 29 -- June 2, 2023}
{London, United Kingdom}{A.~Ricci, W.~Yeoh, N.~Agmon, B.~An (eds.)}
\copyrightyear{2023}
\acmYear{2023}
\acmDOI{}
\acmPrice{}
\acmISBN{}

\acmSubmissionID{83}

\title[AAMAS-2023 Formatting Instructions]{\texttt{PECAN}: Leveraging Policy Ensemble for Context-Aware\\ Zero-Shot Human-AI Coordination}

\author{Xingzhou Lou}
\affiliation{
  \institution{Institute of Automation, Chinese Academy of Sciences}
    \city{Beijing}
  \country{China}
}
\email{louxingzhou2020@ia.ac.cn}
\authornote{Work done while visiting King's College London.}
\authornote{Also with School of Artificial Intelligence, University of Chinese Academy of Sciences, Beijing, China.}

\author{Jiaxian Guo}
\affiliation{
  \institution{The University of Sydney}
  \city{Sydney}
  \country{Australia}}
\email{jguo5934@uni.sydney.edu.au}

\author{Junge Zhang}
\affiliation{
  \institution{Institute of Automation, Chinese Academy of Sciences}
    \city{Beijing}
  \country{China}
}
\email{jgzhang@nlpr.ia.ac.cn}
\authornotemark[2]
\authornote{Correspondence.}

\author{Jun Wang}
\affiliation{
  \institution{University College London}
  \city{London}
  \country{United Kingdom}}
\email{j.wang@cs.ucl.ac.uk}

\author{Kaiqi Huang}
\affiliation{
  \institution{Institute of Automation, Chinese Academy of Sciences}
    \city{Beijing}
  \country{China}
}
\email{kqhuang@nlpr.ia.ac.cn}
\authornotemark[2]

\author{Yali Du}
\affiliation{
  \institution{King's College London}
  \city{London}
  \country{United Kingdom}}
\email{yali.du@kcl.ac.uk}
\authornotemark[3]
\input{0-abstract.tex}


\newcommand{\BibTeX}{\rm B\kern-.05em{\sc i\kern-.025em b}\kern-.08em\TeX}

\begin{document}


\pagestyle{fancy}
\fancyhead{}


\maketitle 

\input{1-introduction.tex}

\input{2-related.tex}

\input{3-method.tex}

\input{4-experiment.tex}

\input{5-conclusions.tex}








\bibliographystyle{ACM-Reference-Format} 
\bibliography{sample}
\balance

\input{supp}


\end{document}

%% file: 0-abstract.tex
\begin{abstract}
Zero-shot human-AI coordination holds the promise of collaborating with humans without human data. Prevailing methods try to train the ego agent with a population of partners via self-play. However, these methods suffer from two problems: 1) The diversity of a population with finite partners is limited, thereby limiting the capacity of the trained ego agent to collaborate with a novel human; 2) Current methods only provide a common best response for every partner in the population, which may result in poor zero-shot coordination performance with a novel partner or humans. To address these issues, we first propose the policy ensemble method to increase the diversity of partners in the population, and then develop a context-aware method enabling the ego agent to analyze and identify the partner's potential policy primitives so that it can take different actions accordingly. In this way, the ego agent is able to learn more universal cooperative behaviors for collaborating with diverse partners. We conduct experiments on the Overcooked environment, and evaluate the zero-shot human-AI coordination performance of our method with both behavior-cloned human proxies and real humans. The results demonstrate that our method significantly increases the diversity of partners and enables ego agents to learn more diverse behaviors than baselines, thus achieving state-of-the-art performance in all scenarios. We also open-source a human-AI coordination study framework on the Overcooked for the convenience of future studies. Codes and demo videos are available at \href{https://sites.google.com/view/pecan-overcooked}{https://sites.google.com/view/pecan-overcooked}.
\end{abstract}

%% file: 1-introduction.tex
\section{Introduction}

Reinforcement learning (RL) has shown remarkable success in various domains, such as gaming AI \cite{silver2017mastering,vinyals2019alphastar,du2019liir,han2019grid,shi2023stay}, robotic manipulation \cite{Liu2022MRN,fang2019curriculum}, traffic control \cite{du2021flowcomm,du2022scalable}, etc.
However, a significant challenge remains in constructing agents that can collaborate effectively with unseen partners, which is especially important under human-AI coordination.
Many real-world applications of human-AI coordination, such as cooperative games \cite{gohst2016overcooked}, self-driving vehicles \cite{resnick2018vehicle,mariani2021coordination} and AI assistants \cite{kakish2019open,andrychowicz2020learning}, can be modeled as zero-shot human-AI coordination tasks. By avoiding the expensive human data collection and human involvement during training, zero-shot human-AI coordination holds the promise of more accessible AI systems that can enhance human capabilities. 
In this approach, an ego agent is trained with partner agents and later interacts with human proxy models or real humans.
Existing methods mainly vary in how the partner agents are acquired and how the ego agent is trained. Self-play methods tried to train ego agents through self play \cite{silver2017mastering,silver2018general,brown2018superhuman,brown2019superhuman}, in which the ego agent is trained to collaborate with a copy of itself. However, this approach has been shown to result in over-fitting to a single cooperative pattern  \cite{hu2020other,lupu2021trajectory} and poor generalization to real human collaborations.
To address this issue, population-based training (PBT) \cite{strouse2021collaborating,zhao2021maximum,lupu2021trajectory} has been employed, where a population of diverse partner agents is used to train the ego agent. The variety of behaviors exhibited by these diverse partners can prevent over-fitting to a single cooperative pattern and improve generalization ability of the ego agent when cooperating with real humans.

However, there are still two limitations of current PBT methods: 1) The finite number of partners in the population restricts the behavioral diversity, making it difficult for the ego agent to coordinate with new partners, particularly those with unique behaviors. Although increasing the population size can address this issue, it requires significant computational resources and decreases the learning efficiency of the ego agent. 2) The ego agent learns a common best response (BR) for every partner, regardless of their behavior patterns. This partner-specific common BR can lead to unsatisfactory human-AI coordination performance as the ego agent lacks the ability to adapt its policy based on the partner's type and behavior pattern.
To address these issues, we propose \textbf{P}olicy \textbf{E}nsemble \textbf{C}ontext-\textbf{A}ware zero-shot human-AI coordinatio\textbf{N} (\texttt{PECAN}), where the policy ensemble method is proposed to increase the diversity of partners without increasing the population size, and the context-aware module is proposed to identify whether the partner is good or poor at the given task, \emph{i.e.} the level of coordination skills. Thus, the ego agent is able to learn level-based common BR rather than common BR for specific partners in the population, which allows the ego agent to acquire more universal coordination behaviors and better coordinate with novel partners.

Specifically, the proposed policy ensemble can generate a new partner whose policy is the weighted average of policy primitives in the population. Since the weights are randomly generated, the policy-ensemble partners are distinct in each iteration. This increases partner diversity and improves the ego agent's ability to collaborate with unseen partners. Additionally, we find that partners created by mixing policies from the same level display better behavioral diversity than those created from the entire population (see Fig. \ref{components} and section \ref{effect_grouping}).
The context-aware module in PECAN is designed to identify the partner's level of coordination skills based on past trajectories. This is achieved through supervised learning. The training data is collected by rolling out various policy ensembles and assigning the corresponding levels of the ensembles' policy primitives as labels. During evaluation, the ego agent updates its recognized context at the start of each episode based on the past trajectory and uses this context to condition its actions.

Our proposed approach, PECAN, presents three main contributions to zero-shot human-AI coordination. 1) PECAN trains generalizable agents without relying on human data. 2) The use of level-based policy-ensemble partners and a context-aware module enhances population diversity without increasing the population size, allowing the ego agent to learn level-based common best responses. 3) PECAN achieves superior performance compared to state-of-the-art baselines in the Overcooked environment \cite{gohst2016overcooked}, as demonstrated by our experimental results and additional studies.



%% file: 2-related.tex
\section{Related Work}
\textbf{Zero-shot Coordination} Zero-shot coordination (ZSC) has been studied in multiple previous studies \cite{cui2021k,treutlein2021new,ribeiro2022assisting}. In ZSC framework introduced by \cite{hu2020other}, two independently trained agents are paired together to fulfill a common purpose in a cooperative game. The paired agents will never encounter each other during training. Thus, the agents must employ compatible policies and should not over-fit to any arbitrary partners or cooperative patterns. Training the agents with diverse partners is effective to alleviate over-fitting to specific partners and improve ZSC performance. Population-based training (PBT) methods \cite{strouse2021collaborating,lupu2021trajectory,zhao2021maximum} have achieved state-of-the-art performance in ZSC. In Fig. \ref{struc}(b), by maintaining a diverse population of training partners, the ego agent in PBT is able to collaborate with a diverse set of partners. FCP \cite{strouse2021collaborating} trains a diverse population by setting different random seeds and including partners of level of cooperation skills and architectures. TrajeDi \cite{lupu2021trajectory} and MEP \cite{zhao2021maximum} adopt explicit diversity objective to generate diverse policies as partners and achieve state-of-the-art ZSC performance. Our PECAN also maintains a population of policies. But the population is used to provide policy primitives for the policy ensemble rather than partners for the ego agent.

\noindent\textbf{Human-AI Coordination} Much previous work in human-AI coordination focuses on planning and learning with human models \cite{carroll2019utility,sadigh2016planning,nikolaidis2013human,kazantzidis2022train}. However, human-AI coordination can be naturally modelled as ZSC tasks, because humans are usually not involved in the training process. And humans usually prefer adaptive AI partners \cite{strouse2021collaborating}. Thus, an adaptive agent with strong ZSC performance is more likely to succeed in human-AI coordination tasks \cite{strouse2021collaborating,zhao2021maximum}.

\noindent\textbf{Diversity in RL} Diversity is a widely discussed issue in RL. SAC \cite{haarnoja2018soft} encourages maximum entropy over the action distributions to improve a single policy's diversity. \cite{hong2018diversity} maximizes the KL divergence between the current policy and some recent policy to encourage diverse action choices. Instead of the diversity of a single policy, many works also focus on diversity of a policy group. \cite{marcolino2013multi} shows that a team of weak yet diverse agents can even defeat teams of strong but uniform agents under certain conditions. \cite{derek2021adaptable} proposed to use generative model to generate diverse policies. MEP \cite{zhao2021maximum} maximizes the population entropy to encourage diversity among a group of policies. EPPO \cite{yang2022towards} uses mean inner product as diversity enhancement regularization and obtains a mutually distinct policy group. Besides diversity over action distributions, multiple previous studies also focus on the diversity over the induced trajectories. DIPG \cite{masood2019diversity} adopts maximum mean discrepancy (MMD) of the induced trajectories as the metric for diversity among policies and encourage qualitatively distinct behaviors. Other measures of diversity such as Jensen-Shannon Divergence \cite{lupu2021trajectory} and mutual information \cite{chenghao2021celebrating} are also adopted to improve diversity among policies and agents in (multi-agent) reinforcement learning. The partner diversity in PECAN is provided by 1) diversity enhancement regularization when generating policy primitives; 2) random selection as well as the random weights of policy primitives when generating policy-ensemble partners.

\noindent\textbf{Policy Ensemble} Policy ensemble (mixture of experts) \cite{jacobs1991adaptive} is the mixture of a group of policy primitives \cite{sutton1999between}, which is able to work individually in the target task. PMOE \cite{ren2021probabilistic} models policy ensemble as a Gaussian Mixture Model (GMM) with learnable weights and practically shows the improved diversity over both action distribution and induced trajectories. EPPO \cite{yang2022towards} uses the arithmetic mean of the policy primitives as the policy ensemble. The learned policy ensemble achieves both high sample-efficiency and strong performance. Instead of learnable weights or arithmetic mean, the weights of policy primitives in PECAN are randomly generated to further improve diversity of the policy ensemble.

%% file: 3-method.tex
\section{Problem Setting}\begin{figure*}[!htb]
    \centering
    \includegraphics[width=0.77\textwidth]{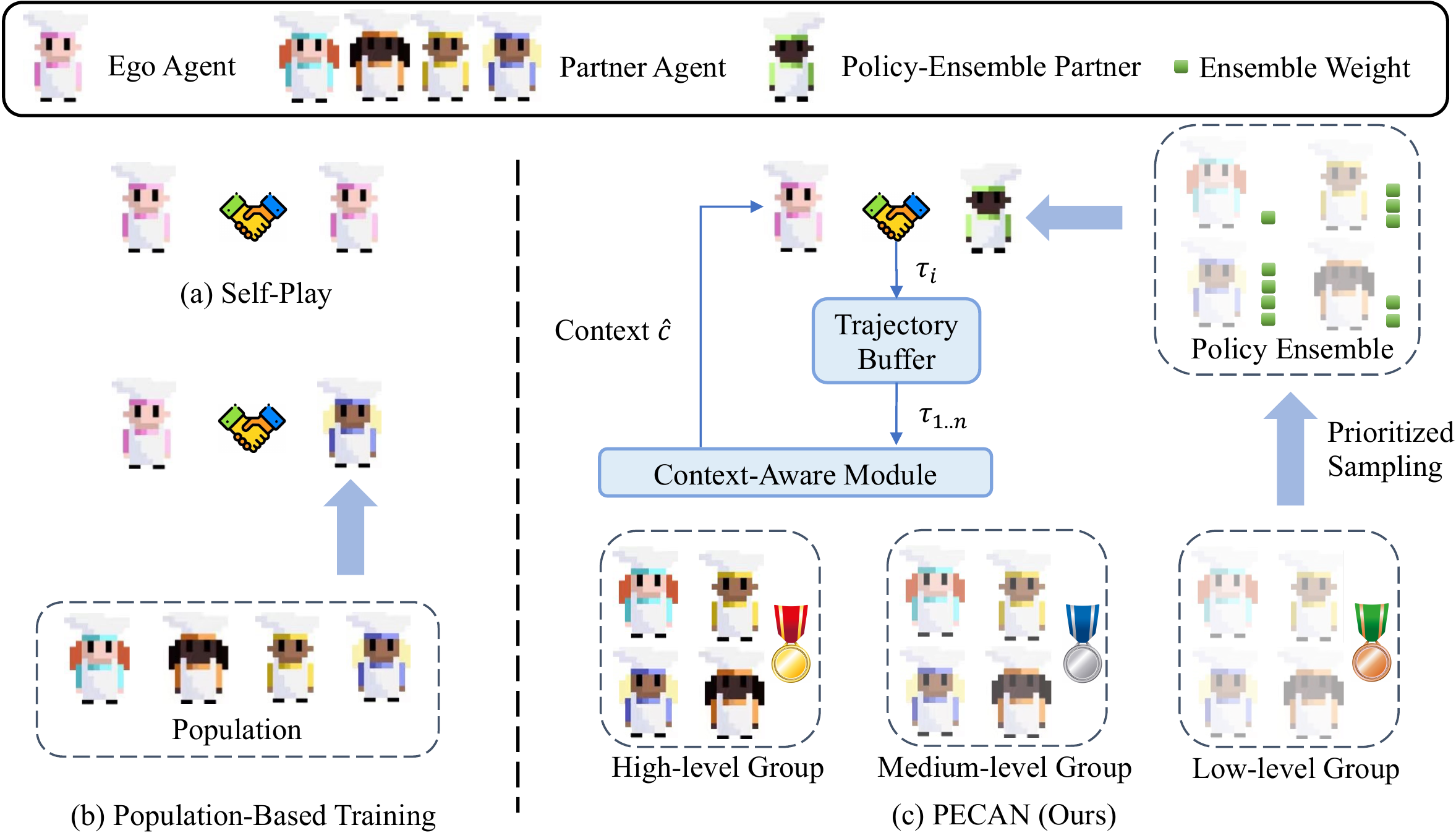}
    \vspace{-0.8em}
    \caption{(a) Self-play training (SP). The ego agent is trained with a copy of itself. (b) Population-based training (PBT). The ego agent is trained with a population of partners. A partner is sampled at each iteration to cooperate with the ego agent. (c) The proposed PECAN method. A different policy-ensemble partner is generated at each iteration. The ego agent will collect trajectories during collaborating with the partner and recognize the partner's level-based context at the beginning of each episode with a pretrained context-aware module introduced in section \ref{context}.}
    \label{struc}
    \vspace{-0.5em}
\end{figure*}
\textbf{Two-player Markov Decision Process} We model the problem as a two-player Markov Decision Process (MDP) $\mathcal{M}=\langle\mathcal{S},\{A^i\},\mathcal{P},\gamma,R\rangle$ \cite{boutilier1996planning}. $\mathcal{S}, \gamma$ are the state space and discount factor. $A^i$ is the action space for the $i$-th agent where $i\in\{1,2\}$. $\mathcal{P}:\mathcal{S}\times A^1\times A^2\times\mathcal{S}\rightarrow[0,1]$ is the transition dynamics. The objective is to maximize the expected reward sum over trajectories $\mathbb{E}_\tau\left[R(\tau)\right]=\mathbb{E}_\tau\left[\sum\limits_t^TR(s,a^1,a^2)\right]$, where $a^1\in A^1, a^2\in A^2$. In our setting, each agent has its own observation and decentralized policy.

\noindent\textbf{Cooperative Tasks} Different from competitive tasks \cite{silver2017mastering,openai2019openai,vinyals2019alphastar} where agents tend to exploit each other, cooperative tasks require agents to collaborate rather than compete and fulfill a shared purpose. In this paper, we consider the problem of common-payoff two-player cooperative games. See the formal representation of two-player cooperative games in \cite{nash1953two} and an intuitive comparison between zero-sum competitive games and common-payoff cooperative games in \cite{carroll2019utility}. MARL can be applied to both competitive \cite{he2016opponent,openai2019openai,silver2017mastering,vinyals2019grandmaster} and cooperative tasks\cite{lowe2017multi,rashid2018qmix,de2020independent,yu2021surprising}. In our cooperative setting, MARL learns a joint-policy $\bm{\pi}=[\pi_1,\pi_2]$ of the two agents so that their expected reward sum over the induced trajectories $\mathbb{E}_{\tau\sim\bm{\pi}}\left[R(\tau)\right]$ is maximized. In PECAN, both the agents in the population and the ego agent are trained by PPO \cite{schulman2017proximal}.


\section{Methodology}
In this section, we will give the details of PECAN. First, we will introduce the population of policy primitives in PECAN. Then, we will introduce the proposed policy-ensemble module and the context-aware module respectively.
\subsection{Population in PECAN}
\label{population}
Like previous PBT methods \cite{zhao2021maximum,strouse2021collaborating,lupu2021trajectory}, PECAN first maintains a population of training partners. To obtain a population, multiple diversity enhancement regularizations have been proposed such as mean inner-product among policies \cite{yang2022towards}, KL divergence among policies \cite{hong2018diversity} and maximum mean discrepancy \cite{masood2019diversity}. In our paper, we directly use the MEP method \cite{zhao2021maximum} to constitute the population considering its computational efficiency and effectiveness.

Specifically, the agent in the population is trained by self-play to learn cooperation ability and maximize the population entropy ($PE$) in \cite{zhao2021maximum}, which quantifies the diversity of the population. The formulation of PE is
\begin{equation}
    PE(\{\pi^1,\pi^2,...,\pi^n\},s_t)=\mathcal{H}\left(\Bar{\pi}(\cdot|s_t)\right)
\end{equation}
where $\Bar{\pi}(\cdot|s_t)=\frac{1}{n}\sum\limits_{i=0}^n\pi^i(\cdot|s_t)$ is the mean policy of the population, and $\mathcal{H}\left(\Bar{\pi}(\cdot|s_t)\right)$ is the entropy of the mean policy.

For agent $i$, the objective of self-play training $J(\pi^i)$ is
\begin{equation}
\label{sp_obj}
    J(\pi^i)=\sum\limits_t\mathbb{E}_{(s_t,\bm{a}_t)\sim\bm{\pi}}\left[R(s_t,\bm{a}_t)+\alpha\mathcal{H}\left(\Bar{\pi}(\cdot|s_t)\right)\right]
\end{equation}
where $\bm{\pi}=[\pi^i,\pi^i]$ is the joint policy, $\bm{a}_t$ is the joint-action sampled from $\bm{\pi}$, $R$ is the reward function and $\alpha$ is the temperature parameter controlling the relative importance of entropy maximization. By maximizing Eq. \ref{sp_obj}, agent $i$ will master high-level cooperation skills, and its policy $\pi^i$ will be encouraged to diversify the current population.

In previous PBT methods \cite{strouse2021collaborating,zhao2021maximum,carroll2019utility}, after the population of training partners is obtained, the ego agent will select a partner from the population for training in each iteration. The partner is selected by either uniform sampling \cite{carroll2019utility,strouse2021collaborating} or prioritized sampling \cite{zhao2021maximum}. However, although the diversity enhancement regularization is adopted, the population diversity is still limited because the population is finite. Moreover, in their training procedure, the ego agent only learns a common BR to every partner in the population, which limits its capacity to coordinate with a novel partner in zero-shot evaluation, since the novel partner is never exposed to the ego agent during training.


\subsection{Diverse Partners with Policy Ensemble}
In this subsection, we will introduce how we adopt policy ensemble to improve the diversity of partners without increasing the population size.  Figure. \ref{struc}(c) gives the idea of our policy ensemble.

\begin{figure*}[ht]
    \centering
    \includegraphics[width=0.83\textwidth]{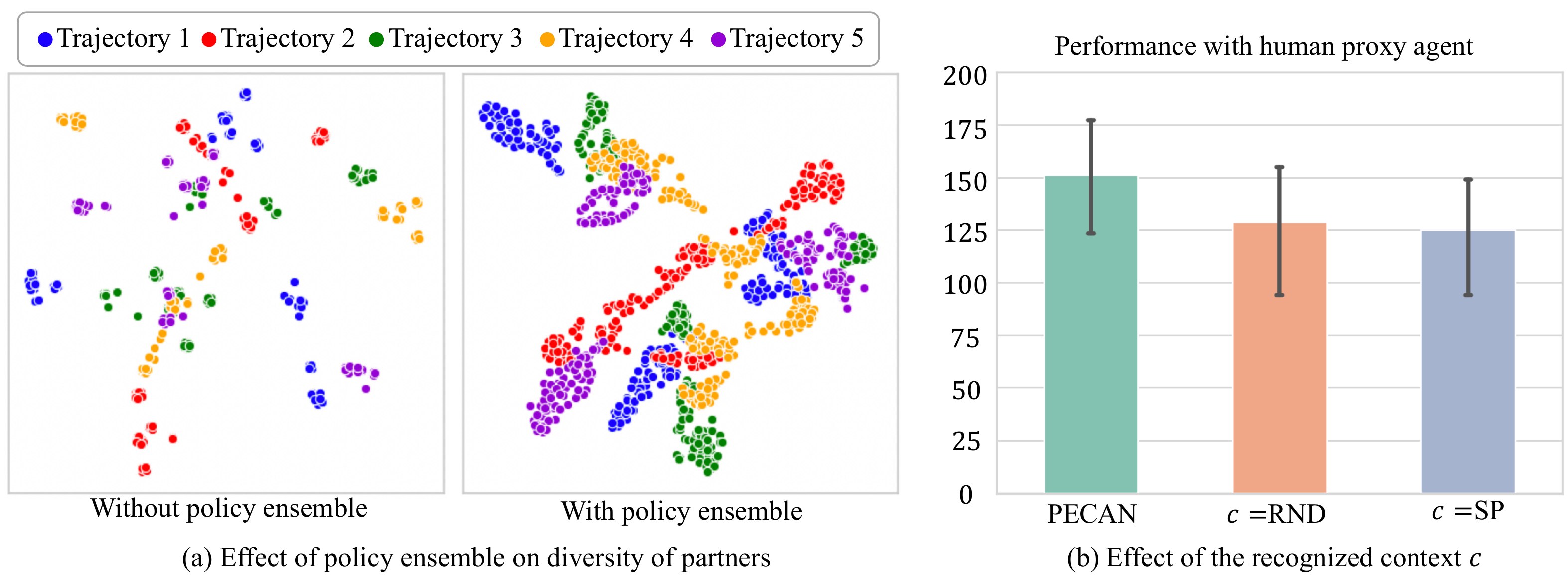}
    \vspace{-0.8em}
    \caption{Effects of each component in PECAN. (a) the t-SNE results of partners' policy $\pi_p(\cdot|s)$ in each episode. The results show policy ensemble is able to improve partners' diversity. (b) $c=RND$ means feeding a random context to the ego agent and $c=SP$ means manually assigning context for the ego agent to indicate that the partner is from the self-play group $G_4$. The results show that only with the correct context, the ego agent can perform well, which means the ego agent's policy is context-aware.}
    \label{components}
\end{figure*}

Instead of directly using policy $\pi^i,\ i\in[1,n]$ in a policy set $G:\{\pi^1,\pi^2,...,\pi^n\}$, 
we use an ensembled policy $\pi_p$ over this set $G$ as the training partner for the ego agents. Specifically, $\pi_{p}$ is the weighted average of policies in $G$, \emph{i.e.},
\begin{equation}
\begin{aligned}
       \pi_p(\cdot|s)=\sum\limits_{\pi^i\in G}\omega_i\pi^i(\cdot|s) \\
       where\ \omega_i\geq0\ and\ \sum\limits_{i=1}^{|G|}\omega_i=1    
\end{aligned}
\end{equation} 
where $\omega_i$ is the weight of $\pi^i$. Intuitively, $\omega_i$ decides the contribution of $\pi^i$ to the policy ensemble $\pi_p$. 
In particular, if the weight of one policy is 1 and the weights of the other policies are 0, the policy ensemble will degenerate into a single policy in the population (as previous PBT method did \cite{zhao2021maximum,strouse2021collaborating,lupu2021trajectory}). In contrast to this special case, we randomly assign different weights to each training iteration, and we observe that the generated partners typically exhibit different policy behaviour than the policy in the population (refer to Fig. \ref{components}). In this way, the partners generated by our method are more diverse than those generated by the vanilla PBT method, even if the population size is the same.



We split the population into three groups $G_1$,$G_2$ and $G_3$ for low, medium and high level of agents according to their self-play performance. Specifically, the checkpoints from different training stages (initial, middle and final) to the population of the corresponding level following \cite{zhao2021maximum,strouse2021collaborating}. For example, the low-level group consists of the initial models from the population, the medium-level group consists of the middle models and the high-level group consists of the final models. And we also provide a self-play group $G_4$ as in \cite{zhao2021maximum}, which only contains a copy of the ego agent itself, to improve the ego agent's self-play performance.

The mixture of policy primitives from the same group tends to remain within the same group, giving partners greater control over their levels. If the ego agent wishes to learn the level-based common BR introduced in a later section, the controllable level of partners is essential. Please refer to section  \ref{effect_grouping} for empirical experiments validating that level-based grouping can provide partners with controllable levels.  It is interesting that we also discovered that level-based grouping can increase more partner diversity . It is consistent with the previous results in \cite{strouse2021collaborating} that level of skills is an important factor of partner diversity.

In each training iteration, one group is chosen to generate a policy ensemble for training the collaboration ability of the ego agent. Instead of randomly sampling the group, we employ a group-level prioritised sampling strategy that selects groups based on the average performance of the ego agent with each group $J(G i)$. This can stabilise the training process and guarantee that the trained ego agent can collaborate effectively with all groups. Specifically, we use rank-based prioritized sampling to assign higher priority to the group with which the ego agent is hard to cooperate as in \cite{vinyals2019grandmaster,zhao2021maximum}. The probability of group $G_i$ being sampled is
\begin{equation}
    p(G_i)=rank\left(\frac{1}{J(G_i)}\right)^\beta \bigg/ \sum\limits_{j=1}^{4}rank\left(\frac{1}{J(G_j)}\right)^\beta
\end{equation}
where $\beta$ is the hyperparameter controlling the strength of prioritization. When $\beta=0$, the prioritized sampling degenerates to uniform sampling as all groups have the same priority. And when $\beta\to\infty$, the group with the worst average performance will be selected with probability 1. As a smooth approximation of the "maximize minimal" paradigm, group-level prioritized sampling helps the ego agent learns to coordinate with partners at the level with the worst coordination performance, and thus can avoid the problem of over-exploiting easy-to-cooperate partners \cite{zhao2021maximum}.

With the randomly assigned weight $\omega$ for each policy primitives and the rank-based prioritized group sampling, the policy-ensemble partners generated in each training iteration are different by design, and thus more diverse than previous methods. To stabilize the learning process, at the beginning of training, the partners have a high probability to be agents directly from the population rather than policy ensembles, during which the ego agent can learn basic cooperation skills. As training goes on, the partners are more likely to be policy ensembles, and the ego agent 
will collaborate with partners that have more diverse behaviours. 
\subsection{Context-Aware Ego Agent}\begin{figure*}[ht]
    \begin{minipage}[t]{0.49\linewidth}
    \centering
    \includegraphics[width=1\textwidth]{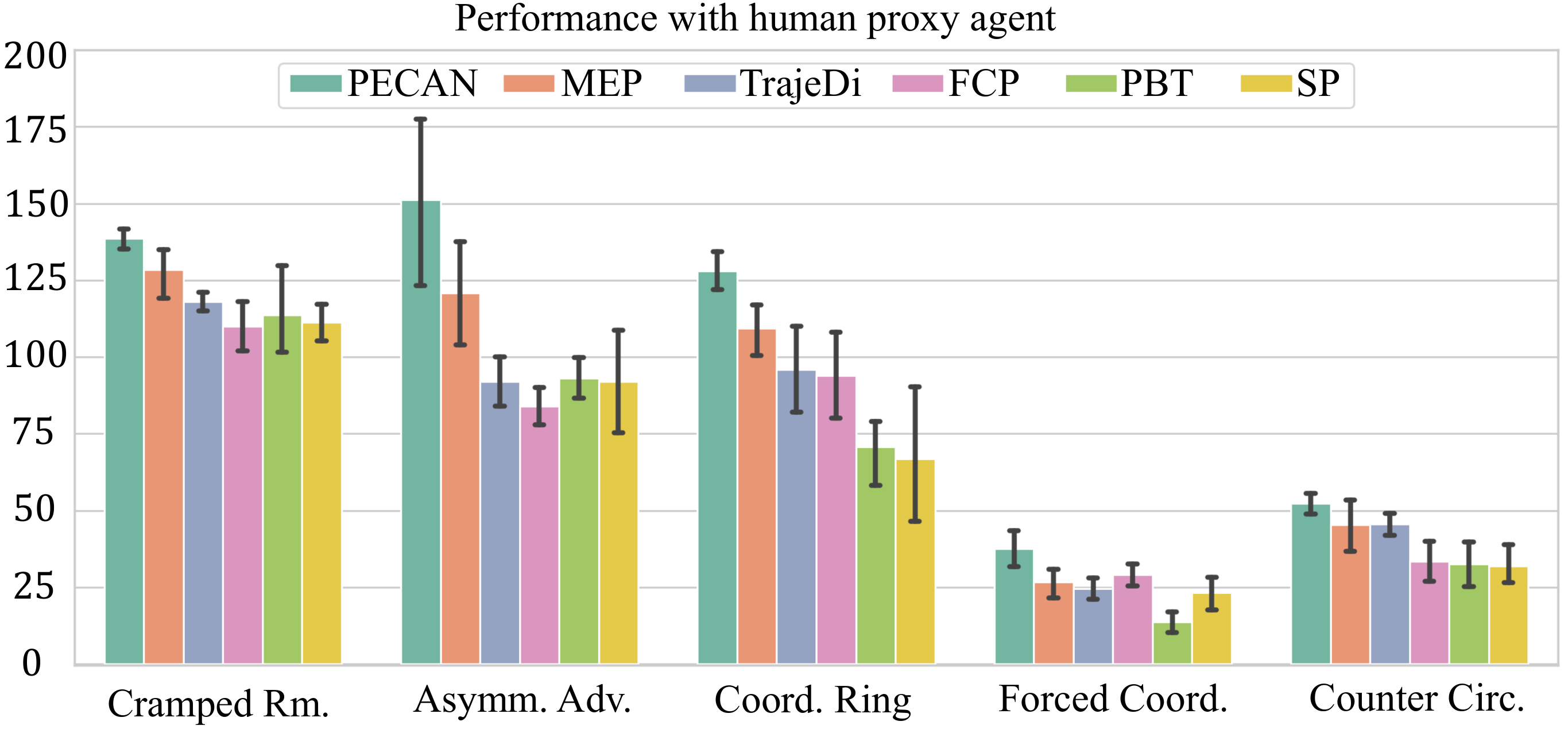}\\
    \centering{\textbf{(a) Overall results}}\\
    \end{minipage}
    \begin{minipage}[t]{0.49\linewidth}
    \centering
    \includegraphics[width=1\textwidth]{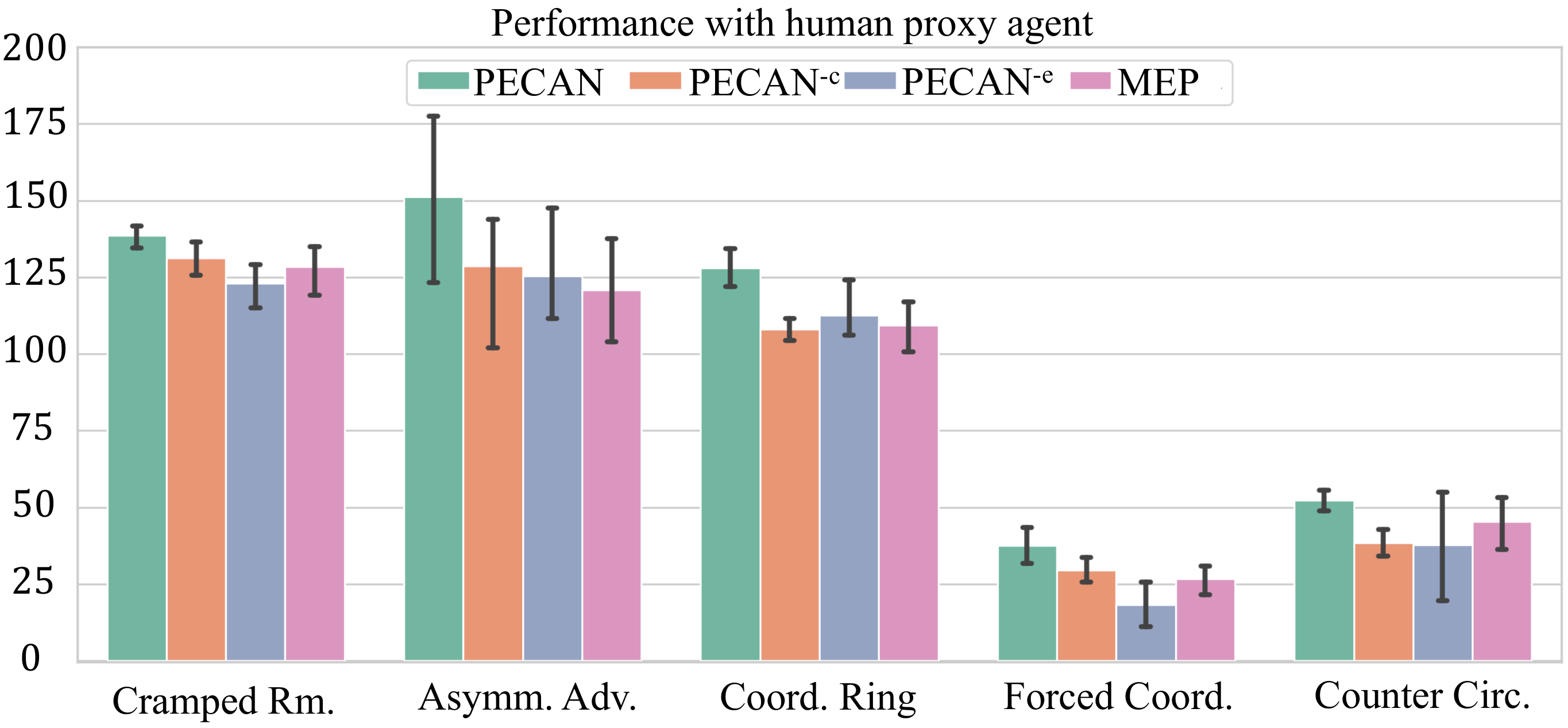}\\
    \centering{\textbf{(b) Ablation study}}\\
    \end{minipage}
    \vspace{-0.8em}
    \caption{(a) Overall results with a human proxy model. PECAN outperforms the baselines on all five layouts. (b) Results of the ablation study. The performance drops when we ablate each component from PECAN, and in some layouts the performance is even worse than the baseline method.}
    \label{main_res}
\end{figure*}\begin{figure}[t]
    \centering
    
    \includegraphics[width=0.49\textwidth]{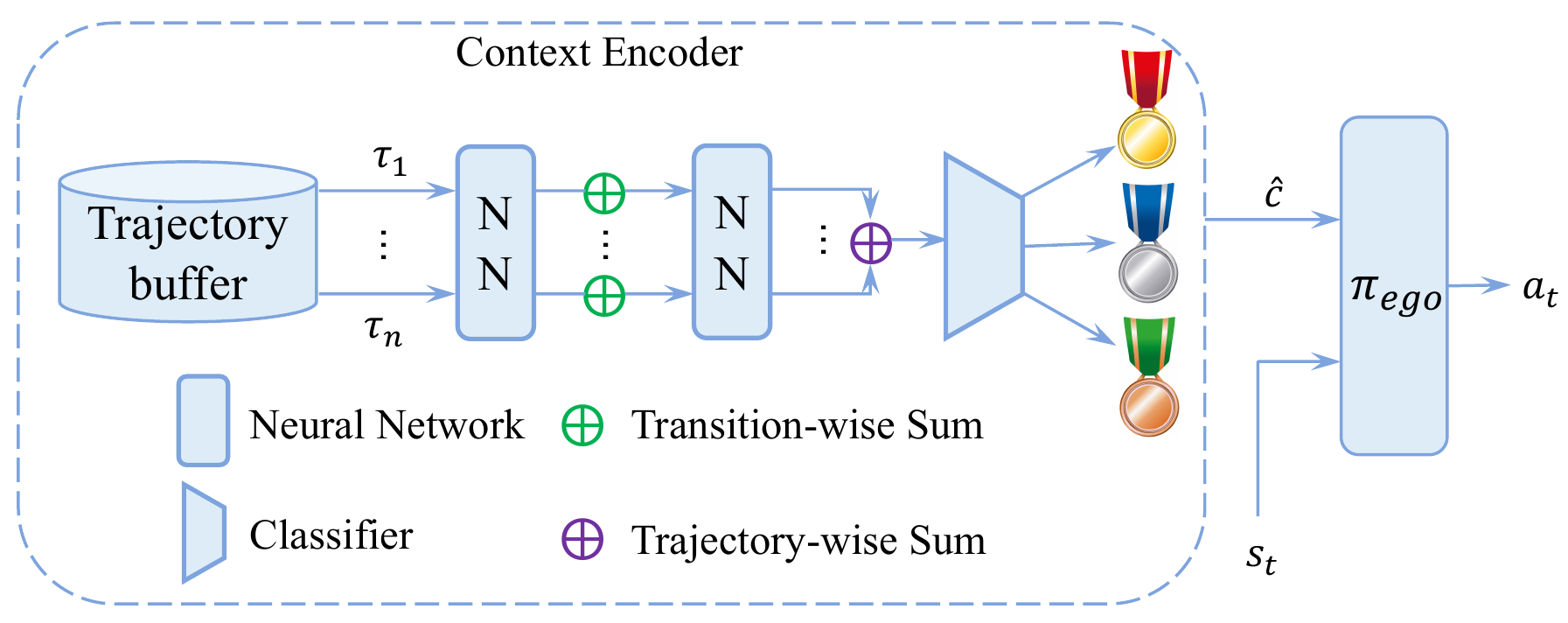}
    \vspace{-2em}
    \caption{The context-aware module predicts the partner's level-based context $\hat{c}$. $\tau_i$ are past trajectories with the same partner from the trajectory buffer. Transition-wise and Trajectory-wise sum are taken respectively to make sure the predictions are irrelevant w.r.t. their orders. The ego agent $\pi_{ego}$ conditions on $\hat{c}$ to make decisions. The context encoder predicts $\hat{c}$ at the beginning of every episode.}
    \label{context_arch}
\end{figure}
\label{context}In current methods, the ego agent only learns a common BR to every partner in the population, which limits its capacity to coordinate with an unfamiliar novel partner. Regardless of diversity of partners, this problem occurs whenever the novel partner in zero-shot evaluation is different from those during training. To alleviate this problem, we propose to learn level-based common BR for the ego agent instead of common BR to specific partners.

In order to learn a level-based BR, we devise a context encoder $f$ to help the ego agent analyze and predict the level of its partner as policy context. Context identification by trajectories has been previously studied \cite{rakelly2019efficient}. But different from their context whose distribution is Gaussian, our contexts are class labels indicating the corresponding group $G_{\{1,2,3\}}$ for low, medium and high level partners, which is determined by their training time as in \cite{strouse2021collaborating}. Therefore, we model the context encoder as a classifier in Fig. \ref{context_arch}.

After the population is obtained, we generate many policy-ensemble partners of different levels and roll out the partners to collect training data $\tau_{\{1,..,N\}}$ and corresponding level-based labels $c_{\{1,..,N\}}$ for the context encoder. The network is updated by gradient descent to predict context $\hat{c}$ with the one-hot label $c$. And the loss function $\mathcal{L}$ is the cross-entropy between $\hat{c}$ and $c$ in Eq. \ref{celoss}.
\begin{equation}
    \label{celoss}
    \mathcal{L}=-\frac{1}{N}\sum\limits_{i=1}^N\sum\limits_{j=1}^3c_{ij}\log\hat{c}_{ij}
\end{equation}
where $N$ is the batchsize. Each input $\bm{\tau}$ in the batch is a set of trajectories with the same label, and $\hat{c}_{ij}=f_j(\bm{\tau}_i)$ is the probability of group $j$ with input $\bm{\tau}_i$.

We predict the context from past trajectories. The sequence of transitions in the trajectory are encoded by a self-attention \cite{vaswani2017attention} module followed by multi-layer perceptrons (MLP). The order of transitions and trajectories is irrelevant to the partner's level. Therefore, we take sum over both encoded transitions within a trajectory and encoded trajectories, so that the prediction result is permutation-invariant w.r.t. transitions in a trajectory and trajectories in the buffer. This architecture has the capacity to represent any permutation-invariant function \cite{zaheer2017deep}. More implementation details of PECAN are given in the supplementary material.

By conditioning on the predicted context $\hat{c}$ indicating the partner's level, the ego agent's policy $\pi_{ego}(\cdot|s,\hat{c})$ learns level-specific coordination skills, which are more universal than previous partner-specific coordination skills. 

The ego agent will learn to coordinate with partners based on their level-based context, and thus reach a level-based common BR to partners of different levels. Different from common BR to specific partners, the level-based common BR is more universal and enables the ego agent to better coordinate with unfamiliar novel partners by analyzing their level of skills and taking actions accordingly.

During zero-shot evaluation with a novel partner, we use the collected trajectories to infer context of the partner. Akin to posterior sampling \cite{strens2000bayesian,osband2013more,rakelly2019efficient}, as more trajectories are collected, the ego agent's belief narrows, and the prediction becomes more accurate. But we would like to note that the process is different from posterior sampling, since we model the context identification as a mapping from trajectories to classes rather than a distribution. Although the context inference uses past trajectories, we do not fine-tune or update any parameters during evaluation. Thus, the evaluation with a novel partner is still zero-shot.

%% file: 4-experiment.tex
\section{Experiments}\begin{figure*}[ht]
    \centering
    \includegraphics[width=0.8\textwidth]{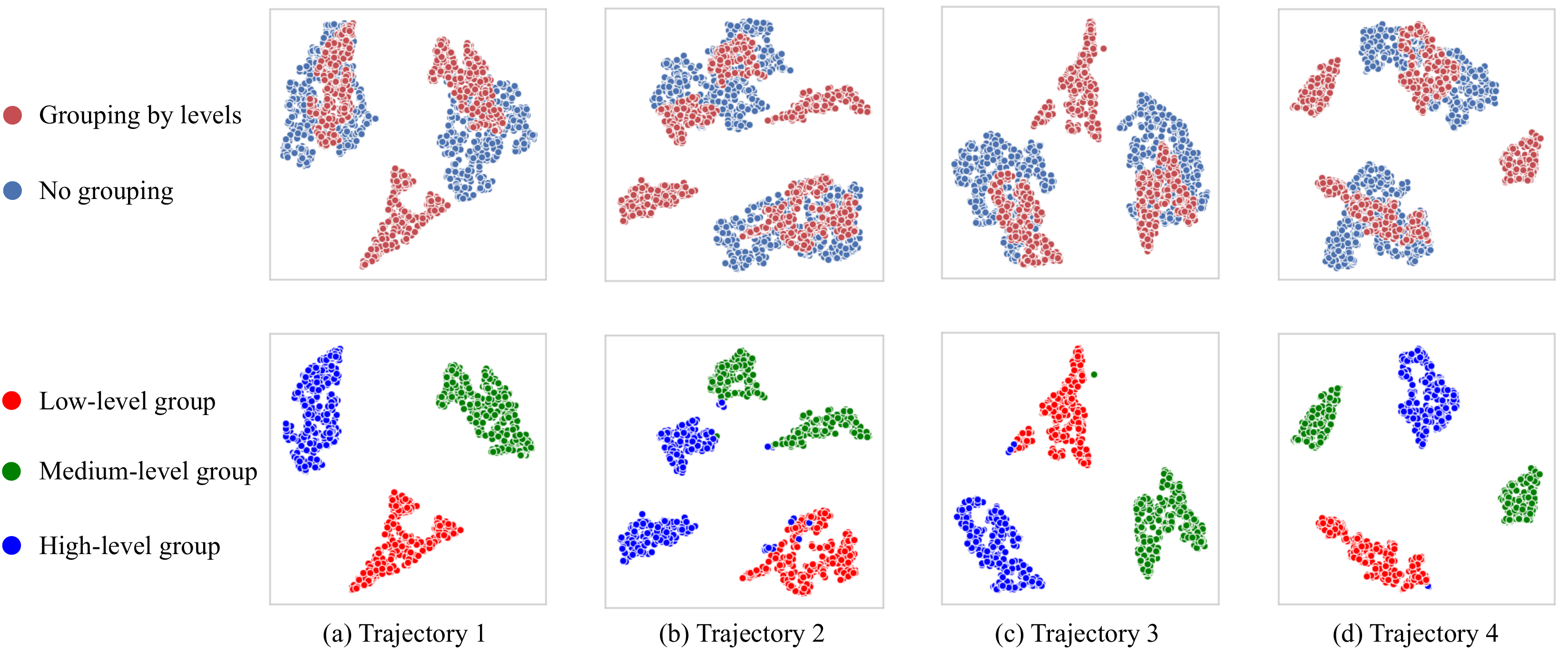}
    \vspace{-0.8em}
    \caption{Comparison between policy ensembles with and without level-based grouping. The ensembles with level-based grouping demonstrate larger diversity (upper row) and their levels of coordination skills are controllable based on the levels of their policy primitives (lower row). Colors in the lower row indicate which group the partner's policy primitives belong to.}
    \label{comparison}
\end{figure*}
In this section, we will first introduce the tasks, baselines and procedures that we adopt to evaluate zero-shot human-AI coordination performance of PECAN. Then, experimental results of coordinating with human proxy models and real human players will be given. Finally, case studies are conducted to show the adaptiveness of PECAN intuitively.

\subsection{Experimental Setting}
\textbf{Tasks} We follow the evaluation protocol proposed in \cite{carroll2019utility} and  evaluate the proposed method on a challenging collaborative game Overcooked \cite{gohst2016overcooked,carroll2019utility}. Five layouts (\emph{Cramped Room, Asymmetric Advantages, Coordination Ring, Forced Coordination} and \emph{Counter Circuit}) in Overcooked are adopted to evaluate the ego agent's ability to coordinate with some novel partners. See more details of the layouts in \cite{carroll2019utility}. Each layout exhibits a unique challenge, which can be overcome if the players coordinate well with each other. The players are required to put three onions in a pot, collect an onion soup from the pot after 20 timesteps and deliver the dish to a counter. The agents will receive 20 points for each dish served. The objective is to serve as many dishes as possible in 1 minute.

\noindent\textbf{Baselines} The baseline methods include Self-play PPO (SP) \cite{carroll2019utility,schulman2017proximal}, population-based training (PBT) \cite{jaderberg2017population,carroll2019utility}, Fictitious Co-Play (FCP) \cite{strouse2021collaborating}, TrajeDi \cite{lupu2021trajectory} and Maximum-Entropy Population-based training (MEP) \cite{zhao2021maximum}.

\noindent\textbf{Procedure} First, we pair the agents with a human proxy model, a behavior-cloning agent that mimics human's behaviors, to test their coordination performance. The effect of each proposed component of PECAN is studied in the ablation study. Besides, we design other experiments to reinforce our claim that (a) policy ensemble is able to improve partners' diversity and (b) the PECAN agent learns a context-aware policy. Then, we recruit human players to evaluate the human-AI coordination ability of PECAN. The human players are required to give their subjective ratings to the agents. Finally, two case studies are conducted to demonstrate the adaptiveness of PECAN in human-AI coordination. More experiment details are given in the supplementary material.
\subsection{Experiments with Human Proxy Model}\begin{figure*}[!ht]
    \centering
    \includegraphics[width=0.8\textwidth]{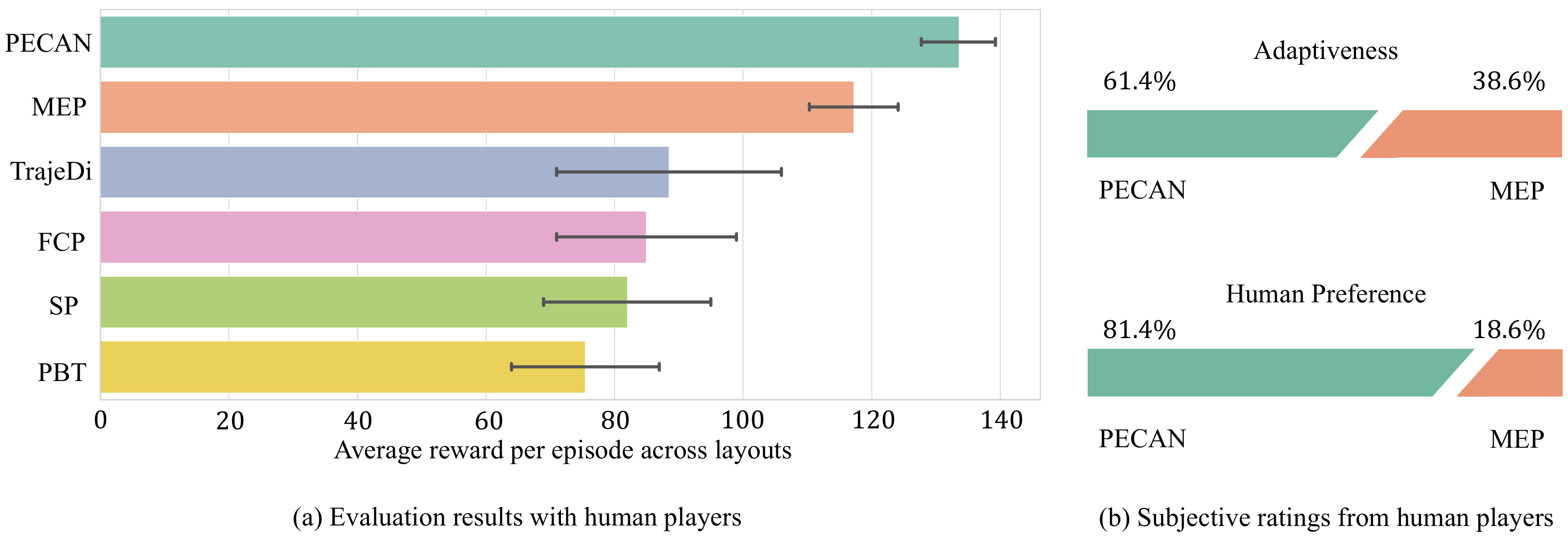}
    \vspace{-0.8em}
    \caption{(a) The coordination performance of each method with real humans. PECAN outperforms the baselines on human-AI coordination. (b) Human players' subjective ratings of the adaptiveness and their preference of the PECAN and the MEP agents. Human players give higher subjective ratings to PECAN on both adaptiveness and personal preferences.}
    \label{human}
\end{figure*}\begin{figure*}[ht]
    \centering
    \includegraphics[width=0.8\textwidth]{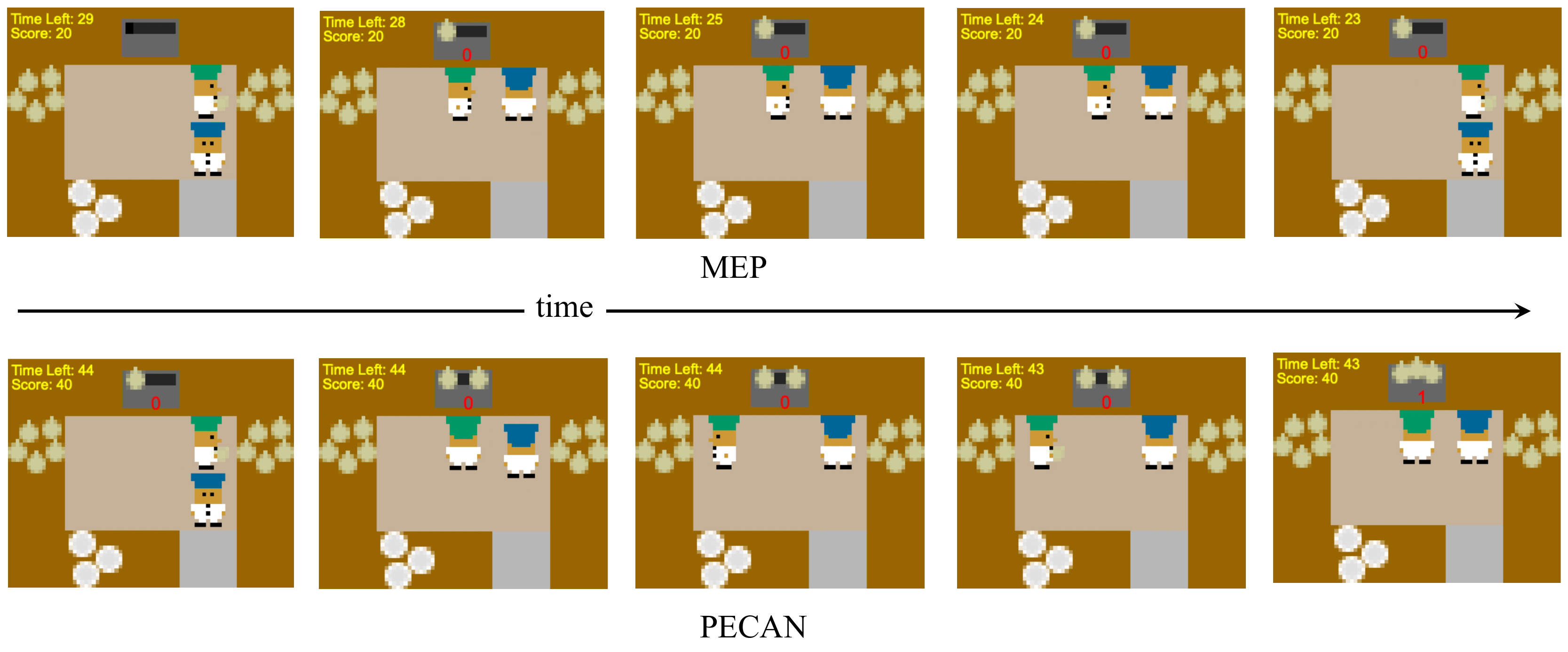}
    \vspace{-0.8em}
    \caption{The MEP agent only learns to pick up onions from the right side. If we intentionally block its way to the onions on the right, it will be stuck and wait until we move away so that it can pick up onions there again. But for the PECAN agent, if we block its way to the onions on the right side, it will turn around to pick up onions on the left with no hesitation. \textcolor{green}{Green} agent is the ego agent and \textcolor{blue}{blue} agent is the human player. }
    \label{case1}
\end{figure*}

\noindent\textbf{Overall Result}
Fig. \ref{main_res}(a) shows the overall coordination performance of PECAN, MEP, SP, PBT, FCP and TrajeDi agents when paired with a human proxy model. We run PECAN and MEP for 4 times with different random seeds and report their average performance. The results of SP, PBT, FCP and TrajeDi are taken from \cite{zhao2021maximum}. For a fair comparison, MEP and PECAN have the same population size. And we adopt the recommended hyperparameter settings for MEP in their paper \cite{zhao2021maximum}. From the results, we can see that PECAN outperforms baseline methods on all five layouts. Especially in Asymm. Adv., the best score by PECAN agents exceeds the baselines by a very large margin (+26.8\%). But in layouts that require less coordination like Cramped Rm., PECAN has relatively marginal performance advantage than the baselines, which indicates that PECAN effectively improves the ego agent's ability to coordinate with its partner rather than to accomplish the task by itself.

\noindent\textbf{Ablation Study}
To study the effect of each component, we ablate policy ensemble and context encoder of PECAN respectively. In PECAN$^{-e}$, policy ensemble is removed, and the partners are chosen the same as in MEP to train the ego agent. In PECAN$^{-c}$, we remove the context encoder and make the ego agent's policy no longer condition on context $c$. 

Fig. \ref{main_res}(b) shows that without policy ensemble and context encoder, PECAN$^{-e}$ and PECAN$^{-c}$ have similar or worse performance than MEP (average performance drop $-18.1$ for PECAN$^{-e}$ and $-14.4$ for PECAN$^{-c}$), while PECAN consistently outperforms the baseline. The result validates the effectiveness of the two proposed modules.

\noindent\textbf{Policy Ensemble and Partner Diversity}
We will empirically validate our previous claim that policy ensemble is able to improve the diversity of partners. We randomly sample some states $s$ from 5 trajectories and partners with/without policy ensemble. Then, we plot the distribution of partners' actions $\pi_p(\cdot|s)$ by t-SNE \cite{van2008visualizing}. 

Fig. \ref{components}(a) gives the visualization results. Each point represents the action distribution $\pi_p(\cdot|s)$ of some partner $p$ over state $s$. There are the same number of data points in the graphs with and without policy ensemble. But because there are limited number of partners without policy ensemble, the points excessively overlap with each other, indicating limited diversity. On the contrary, the action distributions become much more diverse with policy ensemble, which shows that policy ensemble is able to improve partner diversity effectively. The diverse partners allow the ego agent to learn more universal coordination behaviors, and therefore have stronger zero-shot human-AI coordination performance.

\noindent\textbf{Study of the Context-aware Policy}
We claim that the ego agent's policy $\pi_e(\cdot|s,\hat{c})$ is context-aware, which means that the ego agent's policy conditions on context $\hat{c}$. The context encoder in PECAN will recognize the partner's context based on past behaviors and help the ego agent take actions accordingly. If incorrect context is fed to the ego agent, it may make improper decisions. There will be a performance gap between the true context and incorrect context, if the ego agent's policy is context-aware. Thus, we design an experiment to test whether the ego agent's policy is context-aware by manually creating context mismatch.

Specifically, we feed a random or manually-assigned context to the ego agent and pair them with a human proxy model to compare their performance with PECAN, where the context is the recognized context. Fig. \ref{components}(b) gives the results with context mismatch on \emph{Asymmetric Advantages}. $c=RND$ represents replacing the real context with a random context, and $c=SP$ represents forcing the context to indicate the ego agent is collaborating with a partner from the self-play group $G_4$, which is a clear mismatch. The results demonstrates significant performance drop with context mismatch, which means the ego agent's policy is context-aware and the recognized context is effective. And it is worth noting that $c=SP$ has very poor performance. This confirms the conclusions from previous studies \cite{carroll2019utility,strouse2021collaborating} that agents with a self-play cooperative pattern have significantly different behaviors from an agent with strong human-AI coordination performance.

\noindent\textbf{Effect of Level-based Grouping\label{effect_grouping}}
Similar to section 5.1.3, we randomly sample some states $s$ and plot the action distribution $\pi_p(\cdot|s)$ by t-SNE. Fig. \ref{comparison} gives the visualization results. It can been seen that the policy ensembles with level-based grouping are more diverse and their levels are controllable based on the level of their policy primitives, while the policy of partners without grouping always tends to form two clusters, which may be the result of compromise between high-level and low-level partners and limits the diversity of partners. Therefore, the result confirms that the level-based grouping is able to improve partner diversity and provide partners with more controllable levels. And the controllable level of partners allows the ego agent to learn level-based common BR more easily.
\begin{table}[!t]
    \centering
    \caption{The MEP agent has very strong preference to collect dishes from Pot 1 and overlooks dishes in Pot 2, while the PECAN agent shows no such preference and serves much more dishes than the MEP agent.}\label{case2}
    \vspace{-0.8em}
    \begin{tabular}{cccc}
    \toprule
    \multirow{2}{*}[-0.5ex]{Method}      & \multicolumn{2}{c}{Usage}      &    \multirow{2}{*}[-0.5ex]{Dishes served by agent}                    \\
    \cmidrule(lr){2-3}& \ \ \ Pot 1\ \ \ \ \ & Pot 2\\
    \midrule
    PECAN     & 57.1\% & 42.9\%  & 12.25 \\
    MEP     & 88.9\% &  11.1\%  & 6.75 \\
    \bottomrule
  \end{tabular}
  
  \vspace{-0.5em}
\end{table} \subsection{\textbf{Human-AI Coordination}}
We follow the Human-AI coordination test protocol proposed in \cite{carroll2019utility} and recruit 15 human players to participate in the study. We evaluate the average performance across layouts of PECAN and state-of-the-art method MEP and reuse the evaluation results of other baselines in \cite{zhao2021maximum}. The results are compatible because the test procedure is consistent. For the convenience of human-AI coordination experiments on Overcooked, we integrate models from \cite{carroll2019utility} with PantheonRL \cite{sarkar2022pantheonrl}, a newly released library for dynamic training. The code is available here\footnote{\href{https://github.com/LxzGordon/pecan\_human\_AI_coordination}{https://github.com/LxzGordon/pecan\_human\_AI\_coordination}}.

Fig. \ref{human} gives the result of our human study. PECAN outperform all other baselines, and the recruited human players give higher adaptiveness rating to PECAN and noticeably prefer coordinating with PECAN than MEP.

\noindent\textbf{Case Study}
To show the adaptiveness of PECAN, we give two case studies in our human-AI coordination experiments. Demo videos are available at \href{https://sites.google.com/view/pecan-overcooked}{https://sites.google.com/view/pecan-overcooked}.

\noindent\textbf{Case Study 1} See Fig. \ref{case1}. In Cramped Room, we (blue chef) intentionally block the agents'(green chef) way to the onions to see how the agents will react. The MEP agent gets stuck and stand still until we move aside, while our PECAN agent makes adjustments immediately and turn around to pick up onions from the other side. It shows the PECAN agents are more adaptive and capable of adjusting its policy according to the human player's behaviors.

\begin{figure}[!h]
        \vspace{-0.5em}
    \includegraphics[width=0.3\textwidth]{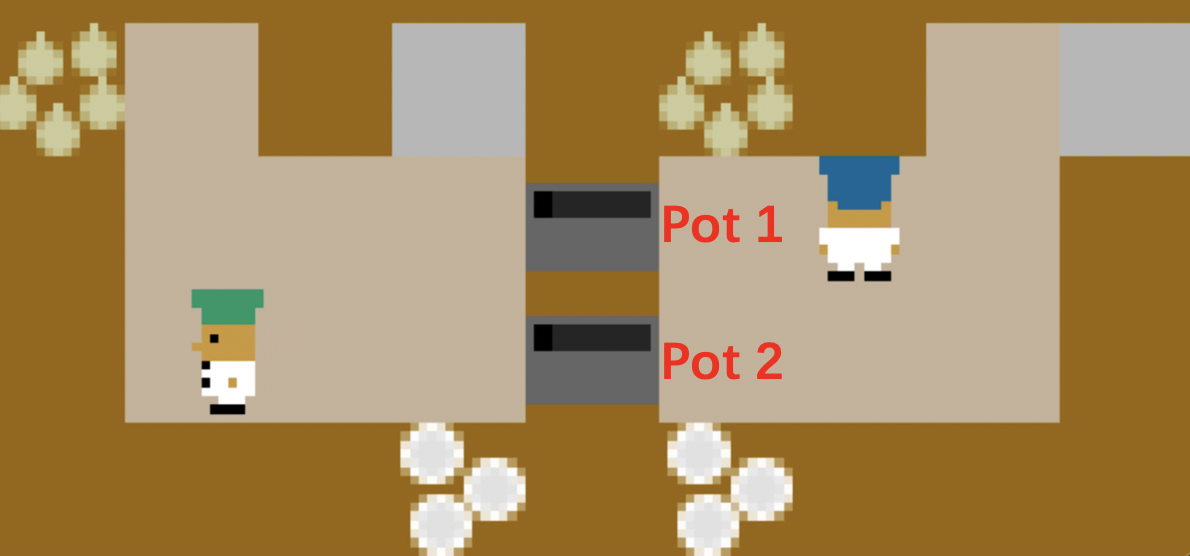}
    \vspace{-0.8em}
    \caption{\emph{Asymmetric Advantages}. AI agents play the chef with \textcolor{green}{green} hat. Pot 1 is 1 step nearer to the serving counter.}
    \label{unident}
    \vspace{-1em}
\end{figure}

\noindent\textbf{Case Study 2} We record the pot usage by the agent in Asymmetric Advantages (see the layout in Fig. \ref{unident}). The trick for the agent (green chef) is to pick up dishes from both pots and serve, because it's much nearer to the serving counter than the human player (blue chef). Table \ref{case2} shows that the PECAN agent has no clear preference over pots than the MEP agent, which has very strong preference for Pot 1. The MEP agent adopts a non-adaptive strategy which has poor generalization to typical human behaviors of using both pots, which further leads to worse performance. However, the PECAN agent's policy is diverse and adaptive, which helps it better coordinate with real humans and serve much more dishes than the MEP agent.

%% file: 5-conclusions.tex
\section{Conclusion and Future Work}
In this paper, we propose a new method (PECAN) for zero-shot human-AI coordination. Policy-ensemble partners and a context encoder are proposed to improve diversity of partners and help the ego agent learn more universal coordination behaviors. We evaluate PECAN with a human proxy model on Overcooked and shows that PECAN is able to outperform all baselines. Ablation studies, further studies and visualization experiments are conducted to demonstrate each component in PECAN. We also organize a human study to evaluate the proposed method's capability of human-AI coordination. The results indicate that PECAN outperforms all other baselines on performance as well as subjective ratings.

Our future work is to study how to analyze and identify the human player's behavior pattern as the ego agents' context (rather than the level-based context in the current method) with the population and no human data during training. In this way, the ego agent can take actions accordingly and better coordinate with humans. This is a very challenging research subject because it requires the ego agent to comprehend human behaviors given only the population of AI agents. And since PECAN is a two-stage method, another future direction is to study how to train it in an end-to-end manner.

%% file: supp.tex
\begin{figure*}[!ht]
    \centering
    \includegraphics[width=0.95\textwidth]{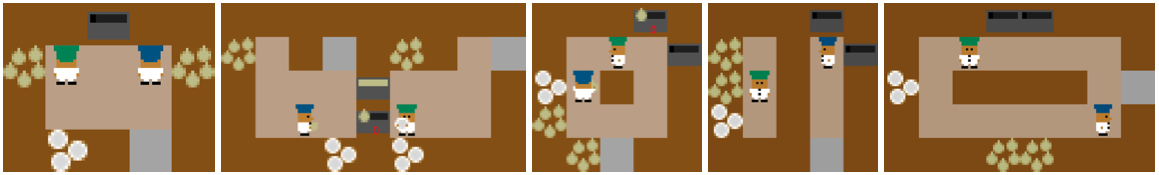}
    \caption{\textbf{Layouts in Overcooked:} from left to right, the layouts are: \emph{Cramped Room, Asymmetric Advantages, Coordination Ring, Forced Coordination and Counter Circuit.}}
    \label{layouts}
\end{figure*}
\newpage

\section*{A. Experiment Details}
\subsection*{A.1 Layouts}
The five layouts in our experiments are given in Fig. \ref{layouts}. There are unique challenges \cite{carroll2019utility} for players to overcome in each layout:

\noindent\textbf{Cramped Room:} This layout presents low-level coordination challenges. In this shared, confined space it is very easy for the agents to collide;

\noindent\textbf{Asymmetric Advantages:} This layout tests whether players can choose high-level strategies that play to their strengths;

\noindent\textbf{Coordination Ring:} In this layout, players must coordinate to travel between the bottom left and top right corners of the layout;

\noindent\textbf{Forced Coordination:} This layout removes collision coordination problems, and forces players to develop a high-level joint strategy, since neither player can serve a dish by themselves.

\noindent\textbf{Counter Circuit:} This layout involves a non-obvious coordination strategy, where onions are passed over the counter to the pot, rather than being carried around.
\subsection*{A.2 Implementation Details}
The population in PECAN is trained with population entropy maximization proposed in \cite{zhao2021maximum}. Thus, we remain the recommended agent architecture and hyperparameter settings for population training, such as 5 initial agents with different random seeds, learning rate of $8\times10^{-4}$ for agent training, reward shaping horizon $5\times10^6$ and so on. As in \cite{strouse2021collaborating}, the initial, the middle and the final checkpoints of the 5 agents are saved to form the population. In our level-based grouping, the initial checkpoints form the low-level group, the middle checkpoints form the middle-level group, and the final checkpoints form the high-level group. $\beta$ for group-level prioritized sampling is $3$ in our experiments. The probability of directly selecting a partner from the population rather than providing a policy-ensemble partner decays to $0.1$ after 50 training iterations.

In the context encoder, each transition vector in the trajectories are first encoded by a fully-connected layer with hidden size 128 . Then, the trajectory sequence are encoded by 5 self-attention \cite{vaswani2017attention} modules with dimensions of $Q,K,V$ matrices being 128. The transition-wise sum is taken after the self-attention modules. Then, 2 fully-connected layers with hidden size 128 follows. Before the last fully-connected layer, the trajectory-wise sum is taken. We adopt the $LeakyReLU$ \cite{maas2013rectifier} activation function for most layers of the network and $Sigmoid$ for the layer before the output layer. The network is optimized by Adam \cite{kingma2014adam} with learning rate of $10^{-3}$. In the cross-entropy loss function of the context encoder, batch size $N=128$ and the input set of trajectories $\bm{\tau}$ randomly contains $1,2$ or 3 trajectories to make sure the ego agent can predict the partner's context from different number of trajectories.
\section*{B. Human Study Statement}
Below is the statement of our human-AI coordination study, which are shown to the participants. The participants are required to sign this statement after reading carefully.

\subsection*{B.1 Purpose}
You have been asked to participate in a research study that studies human-AI coordination. We would like your permission to enroll you as a participant in this research study.

The instruments involved in the experiment are a computer screen and a keyboard. The experimental task consisted of playing 5 layouts of the computer game Overcooked and manipulating the keyboard to coordinate with the AI agent to cook and serve dishes. You will be given specific instructions for the task before it begins.
\subsection*{B.2 Procedure}
In this study, you should read the experimental instructions and ensure that you understand the experimental content. The whole experiment process lasts about 40 minutes, and the experiment is divided into the following steps:

1) Read and sign the experimental statement;

2) Test the experimental instrument, and adjust the seat height, sitting posture, and the distance between your eyes and the screen. Please ensure that you are in a comfortable sitting position during the experiment;

3) You will be paired with a dummy agent in a demo layout. You should comprehend the specific instrument operation rules and be familiar with the experimental process in the demo layout; 

4) Start the formal experiment. You will be paired with an AI agent and play 8 rounds in each layout. Please cooperate with the AI agent to complete the 5 layouts and get as much scores as possible within 1 minutes. Note that after each layout, you should have a rest;

5) After the experiment, you need to fill in a questionnaire.
\subsection*{B.3 Risks and Discomforts}
The only potential risk factor for this experiment is trace electron radiation from the computer. Relevant studies have shown that radiation from computers and related peripherals will not cause harm to the human body.
\subsection*{B.4 Costs}
Each participant who completes the experiment will be paid 100 RMB, and the top 5 participants will be paid another 100 RMB.
\subsection*{B.5 Confidentiality}
The results of this study may be published in an academic journal/book or used for teaching purposes. However, your name or other identifiers will not be used in any publication or teaching materials without your specific permission. In addition, if photographs, audio tapes or videotapes were taken during the study that would identify you, then you must give special permission for their use.
\subsection*{B.6 Participant Declaration}
I confirm that the purpose of the research, the study procedures and the possible risks and discomforts as well as potential benefits that I may experience have been explained to me. All my questions have been satisfactorily answered. I have read this consent form. My signature below indicates my willingness to participate in this study.